\documentclass[12pt]{article}

\usepackage{graphicx}
\usepackage{subfig}
\usepackage{authblk}
\usepackage[font={footnotesize}]{caption}

\usepackage[fontsize=8]{pdfcomment}
\usepackage{color}
\newcommand{\vcns}{virtual nervous system}
\newcommand{\Vvcns}{Virtual nervous system}
\newcommand{\Vcns}{VNS}
\newcommand{\Vcnss}{VNSs}

\title{Virtual Nervous Systems for Self-Assembling Robots --- A preliminary report} 

\author[1]{Nithin Mathews}
\author[2]{Anders~Lyhne~Christensen}
\author[1]{Rehan O'Grady}
\author[1]{Marco Dorigo}
\affil[1]{IRIDIA, Universit\'e Libre de Bruxelles}
\affil[2]{BioMachines Lab, Instituto Universit\'ario de Lisboa (ISCTE-IUL)}

\date{}

\begin{document} 

\maketitle 

\begin{figure}[b!]
\centering
\includegraphics[width=0.65\linewidth]{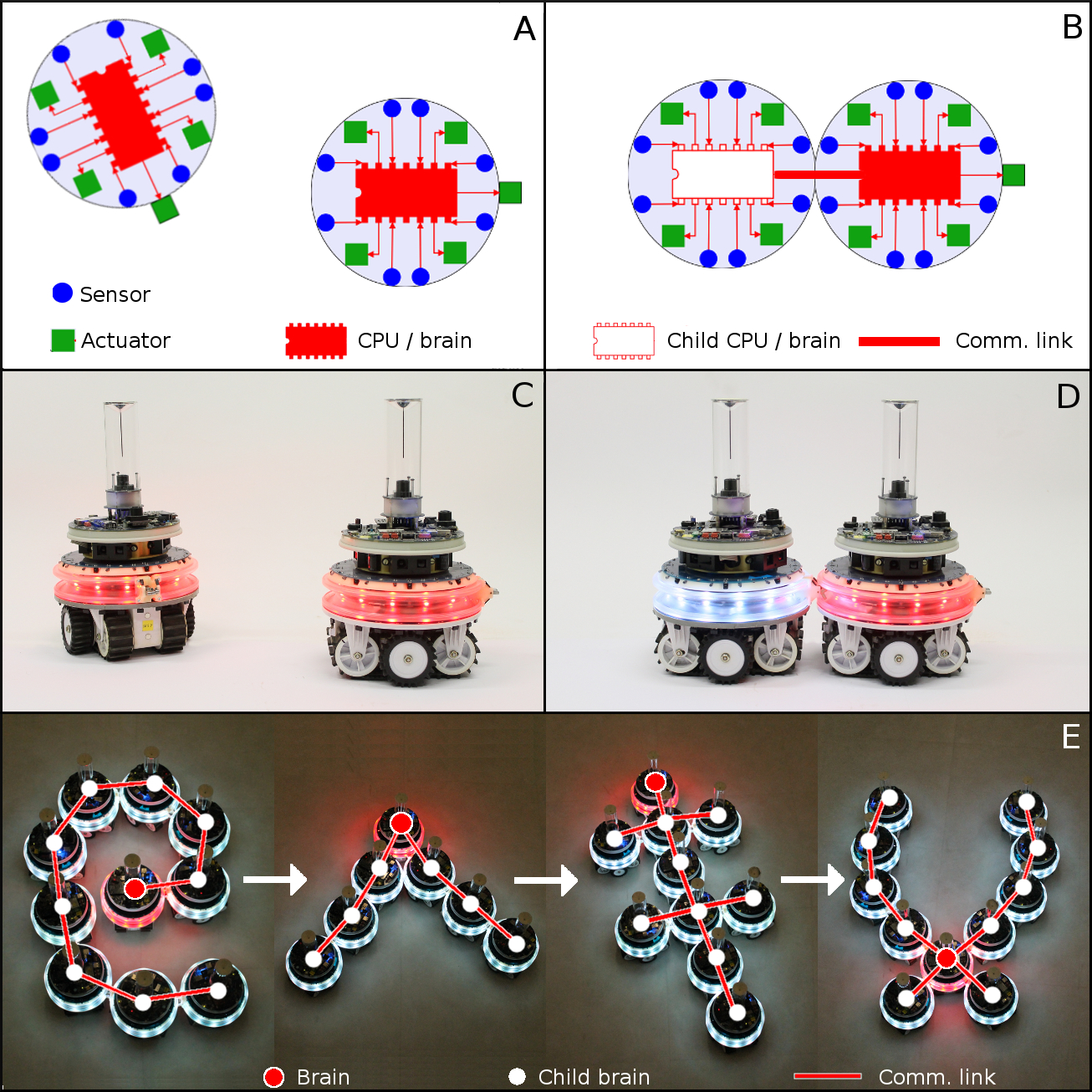}
\caption{\textbf{\Vvcns\ for robots.} Brain robots are illuminated in red.  (\textbf{A}) Schematics of two independent robot bodies. (\textbf{b}) Schematics of the two robots that are now physically connected using a \Vcns\ to form a single entity controlled by a single brain. Through the VNS, the brain robot has access to all sensors and actuators in the composite body. (\textbf{C}) Two independent robots. (\textbf{D}) The two robots autonomously form a physical connection; when the robots attach, they form a communication link: they are now a single entity controlled by the rightmost robot. (\textbf{E}) Snapshots from an experiment in which a series of different robot bodies with VNSs were autonomously formed through repeated self-assembly, disassembly and re-assembly.}
\label{vcns_for_robots}
\end{figure}

We define the nervous system of a robot as the processing unit responsible for controlling the robot body, together with the links between the processing unit and the sensorimotor hardware of the robot --- i.e., the equivalent of the central nervous system in biological organisms. 
Here, we present autonomous robots that can merge their nervous systems when they physically connect to each other, creating a \vcns\ (\Vcns), see Fig.~\ref{vcns_for_robots}. 
We show that robots with a \Vcns\ have capabilities beyond those found in any existing robotic system or biological organism: they can merge into larger bodies with a single brain (i.e., processing unit), split into separate bodies with independent brains, and temporarily acquire sensing and actuating capabilities of specialized peer robots. \Vcns-based robots can also self-heal by removing or replacing malfunctioning body parts, including the brain.

Over the last ten years, we have been developing the basic technologies required to enable \Vcnss. We have developed robots that can form physical connections with each other and the corresponding control algorithms that allow them to do so autonomously ~\cite{ChrOGrDor2007:ram,DorFloGam-etal2013:ram,MatChrOGr-etal2011:iros,MonGamFlo-etal2005:ram,OGrChrDor2009:tr}.  However, these physically connected robots retained independent control of their own bodies, so that the physically connected ensemble could display only crude levels of coordination~\cite{TriDor2006:biocyb}.

\begin{figure}[b!]
\centering
\subfloat{\label{fig:ex1A}\includegraphics[width=0.325\linewidth]{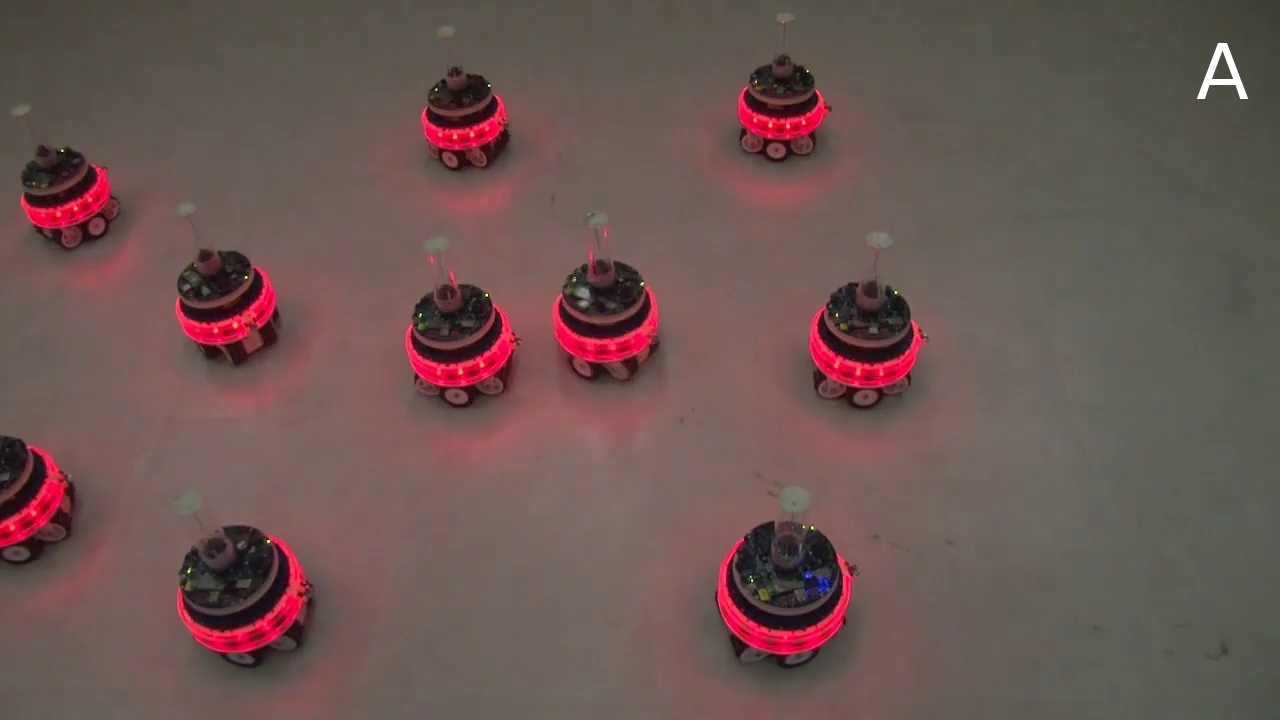}}\hspace{0.005mm}
\subfloat{\label{fig:ex1B}\includegraphics[width=0.325\linewidth]{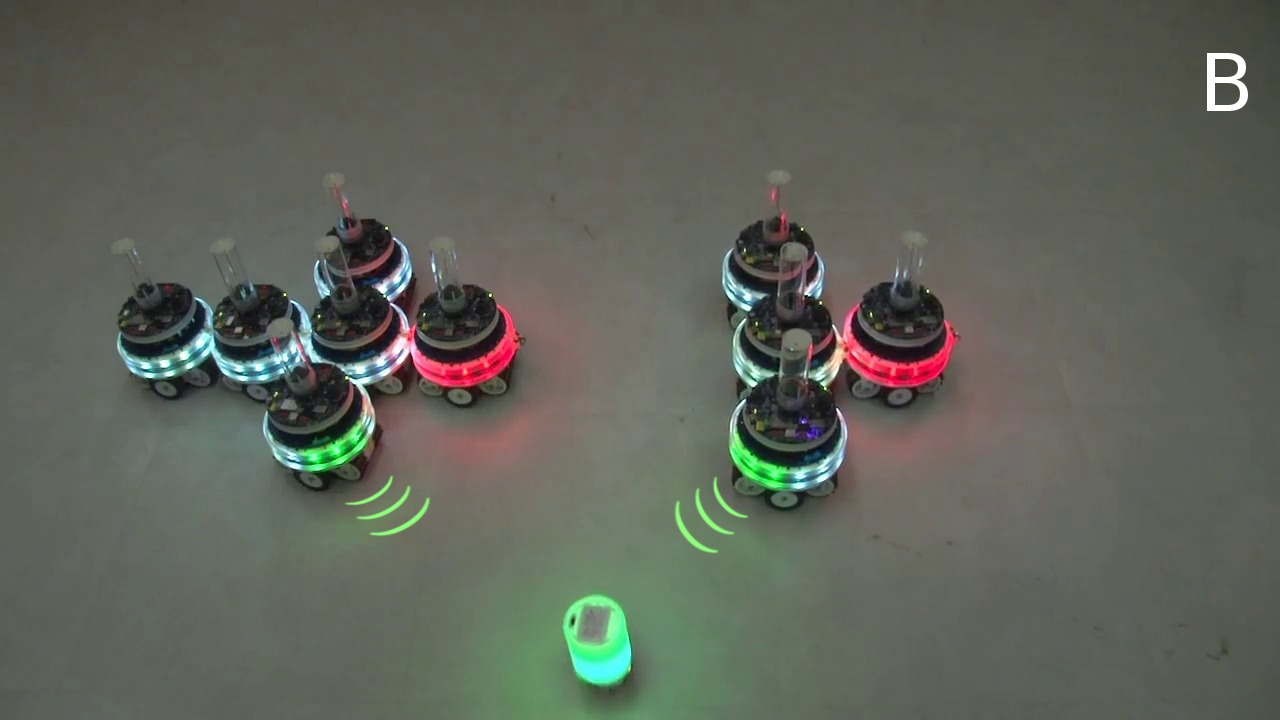}}\hspace{0.005mm}
\subfloat{\label{fig:ex1B_motion}\includegraphics[width=0.325\linewidth]{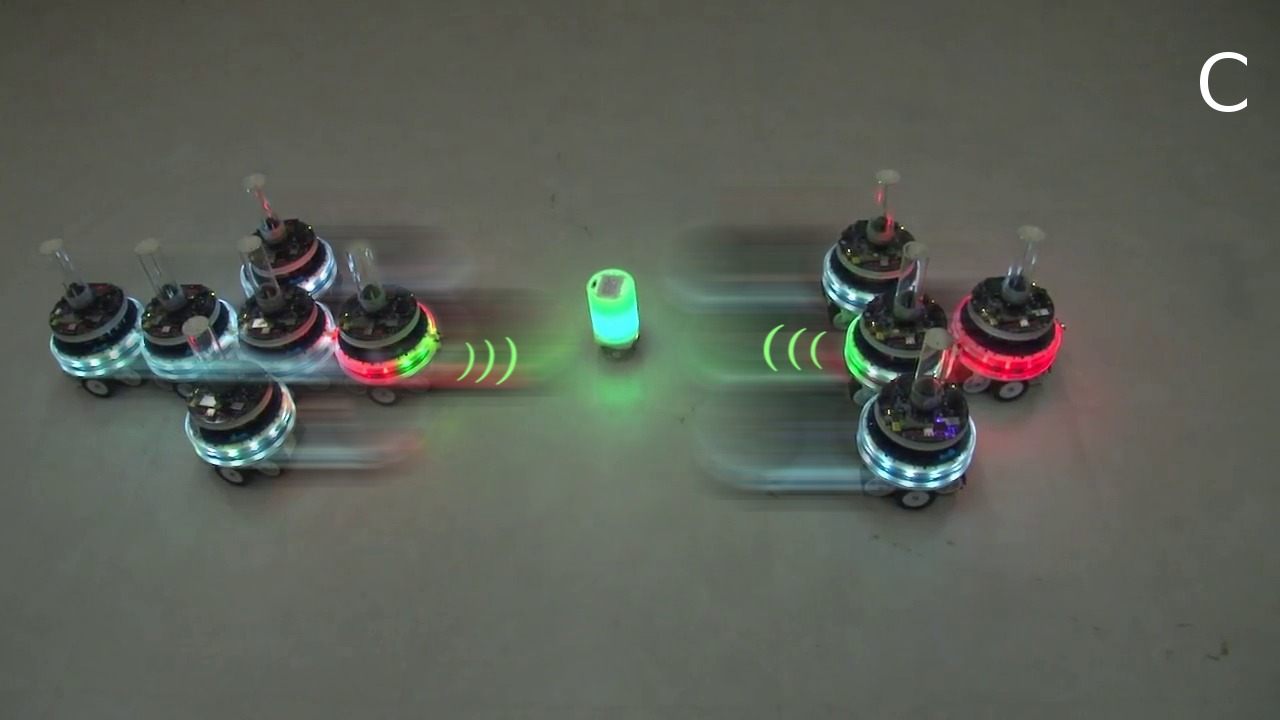}}\\
\vspace{-0.3cm}
\subfloat{\label{fig:ex1C_motion}\includegraphics[width=0.325\linewidth]{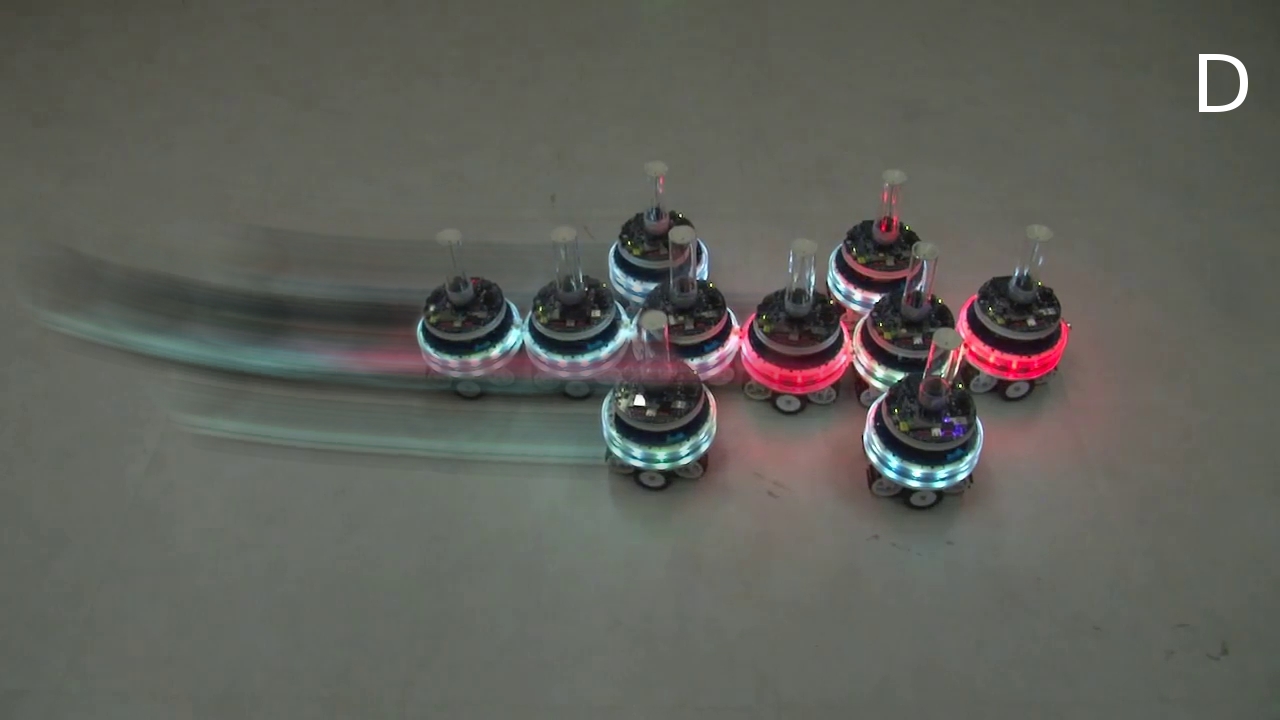}}\hspace{0.005mm}
\subfloat{\label{fig:ex1D}\includegraphics[width=0.325\linewidth]{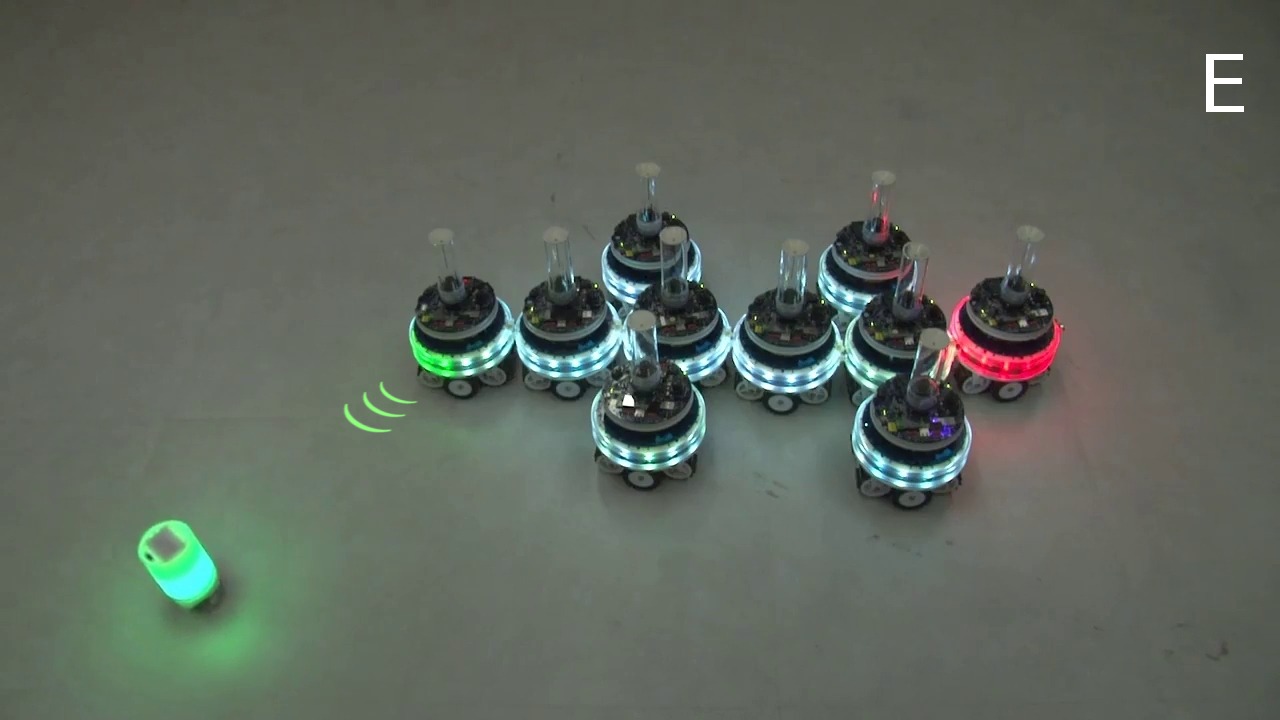}}\hspace{0.005mm}
\subfloat{\label{fig:ex1D_motion}\includegraphics[width=0.325\linewidth]{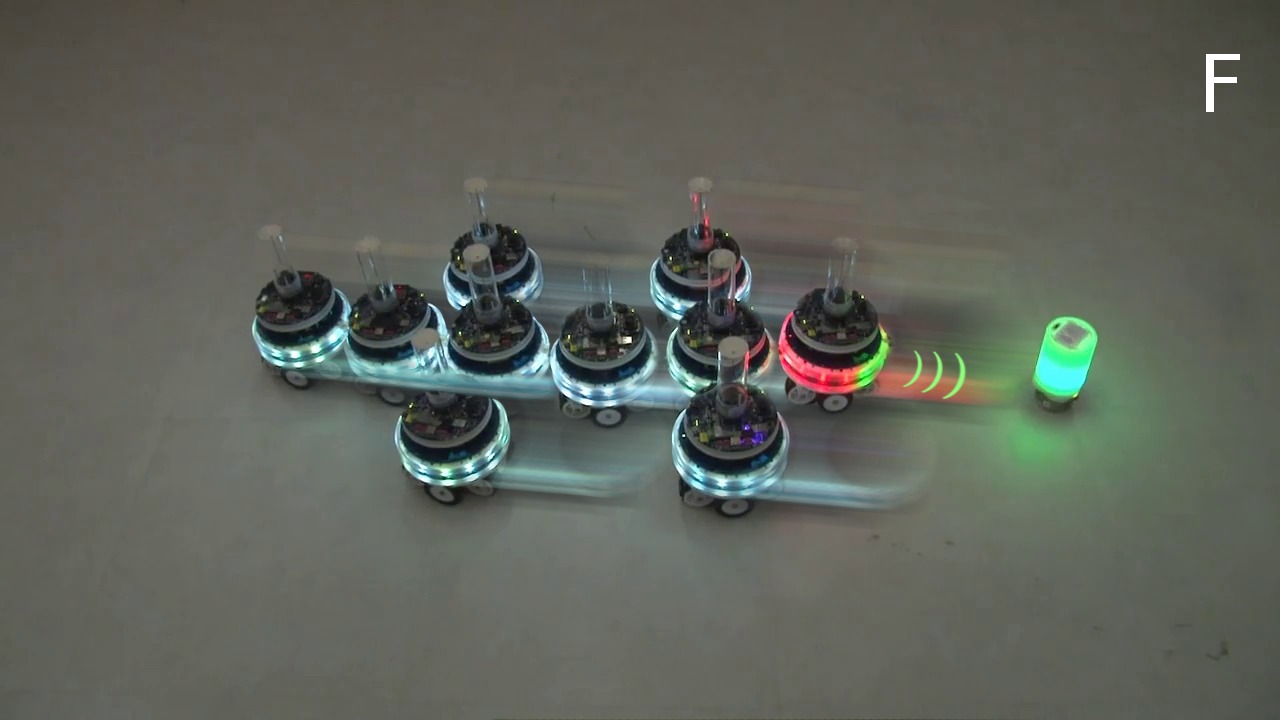}}\\
\caption{\textbf{Reaction to stimuli of self-assembled composite robot bodies with a \Vcns.} Brain robots are illuminated in red. 
(\textbf{A,B}) Ten autonomous robots self-assemble to form two composite robots with \Vcnss. 
(\textbf{B,C}) The two composite robots with \Vcnss\ react to a green stimulus by pointing to the stimulus (lighting up their closest LEDs) and retreating from it when it gets too close.
(\textbf{D}) The two composite robots merge into a single larger composite robot with a single brain. The composite robot on the left cedes authority to the brain of the composite robot on the right which becomes the brain of the newly formed larger composite robot. 
(\textbf{E,F}) The larger composite robot again reacts to the green stimulus by pointing at the stimulus and retreating when it is too close. Note that robot colors are only for illustrative purpose and play no role in the robot control algorithms.}
\label{assembly_and_merge}

\end{figure}

The final step was to give our robots the capability to fuse their robotic nervous systems into a \Vcns, see Fig.\ref{vcns_for_robots}.
Figure~\ref{assembly_and_merge} shows how \Vcnss\ work in a group of ten autonomous robots.  The robots self-assemble into two composite robots~\cite{MatChrOGr-etal2011:iros} that display \Vcns-based behaviors --- each composite robot acts as a single entity, with a single brain making decisions, and the whole composite body reacting as one.
The decisions made by the brain are based on sensory information collated from all of the sensors distributed across the multiple robots of its composite body.
Each composite body reacts to a stimulus by first `pointing' at the stimulus with its LEDs, then retreating from the stimulus. When a composite body points to the stimulus, only the LEDs closest to the stimulus illuminate, independently of the constituent robots to which those LEDs happen to belong. When moving away from the stimulus, movements of all of the different wheel actuators (on all of the constituent robots) are coordinated by the brain robot in the VNS to ensure smooth motion of the entire composite body.

Our \Vcns-based robots have a communication system that allows them to exchange data once they are physically connected.  
When two independent robots come together, each with its own brain, one of the brains needs to cede authority. We addressed this delegation of authority issue by imposing a hierarchical tree-structure on the physical connection topology of the robots and then constraining the logical topology of our \Vcns\ to follow the physical connection topology. Given such a tree-structure there is always a single node that can unambiguously be identified as the root and that can naturally take the role of brain of the \Vcns. The brain of any connecting robot body will cede authority to this root robot. 

To ensure that the brain taking over can quickly obtain knowledge about its new body, each robot maintains recursive knowledge of the connection topology of all of its descendants (children, children's children, etc). By maintaining recursive self-knowledge throughout the hierarchy, we avoid the need for a time consuming self-discovery process~\cite{Bongard-Science-2006} when splitting or merging occurs.
When a split occurs, the root node of the uncoupling body segment already has all the knowledge it needs to become the brain of the newly independent body. When a merge occurs only a single message needs to be passed up the VNS from the brain that is ceding authority to the brain of the new resulting body. The information contained in the message is incrementally updated by each intermediate node with local topological information. 

\begin{figure}[t!]
\centering
\includegraphics[width=0.45\linewidth]{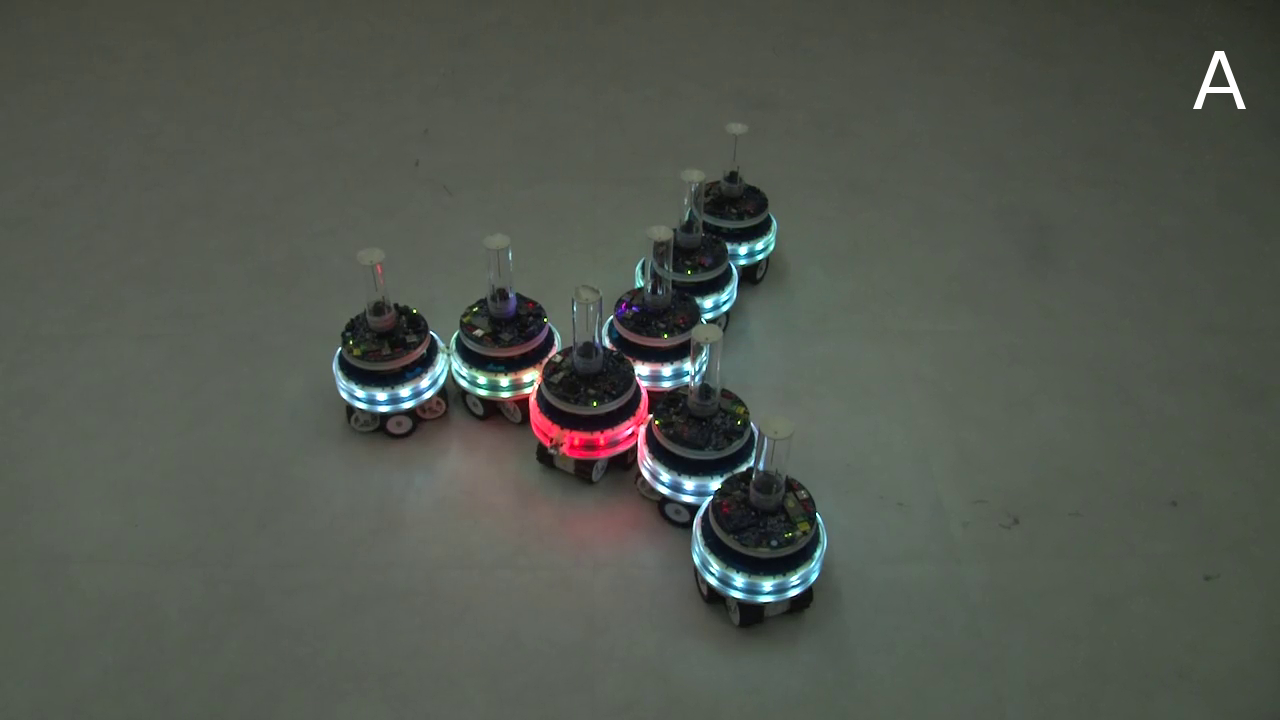}\hspace{0.002mm}
\includegraphics[width=0.45\linewidth]{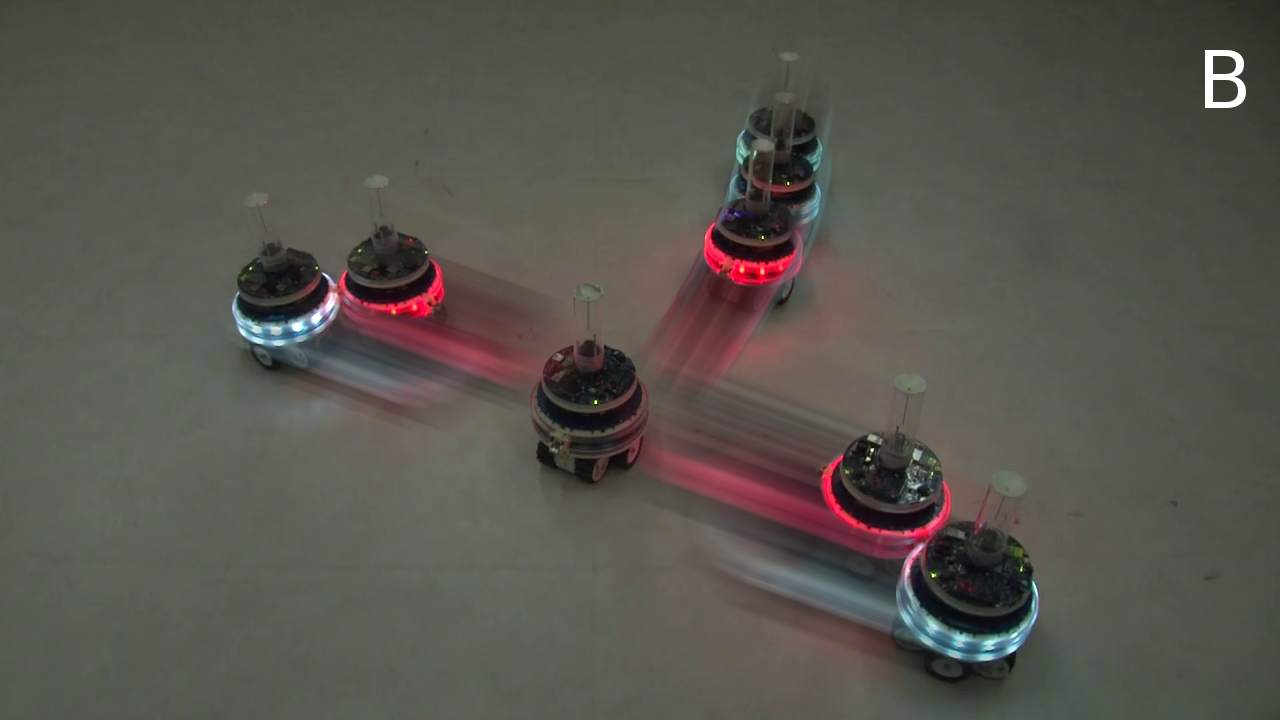}\\
\includegraphics[width=0.45\linewidth]{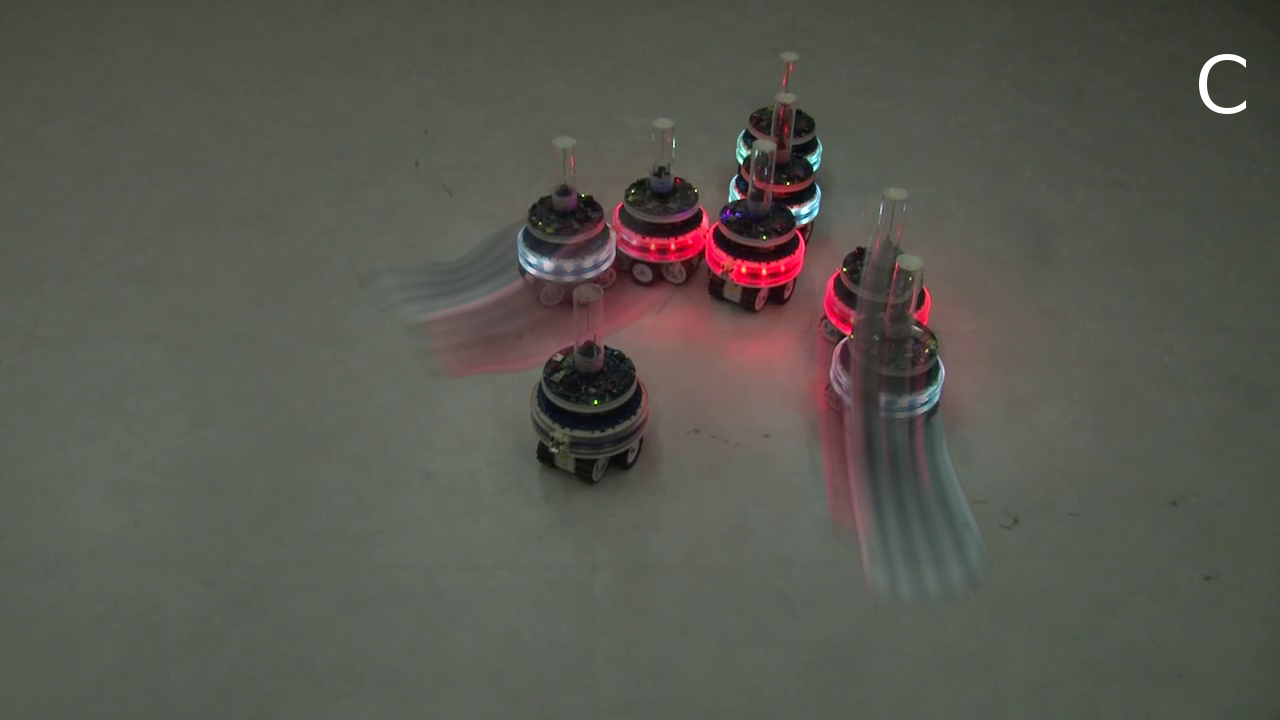}\hspace{0.002mm}
\includegraphics[width=0.45\linewidth]{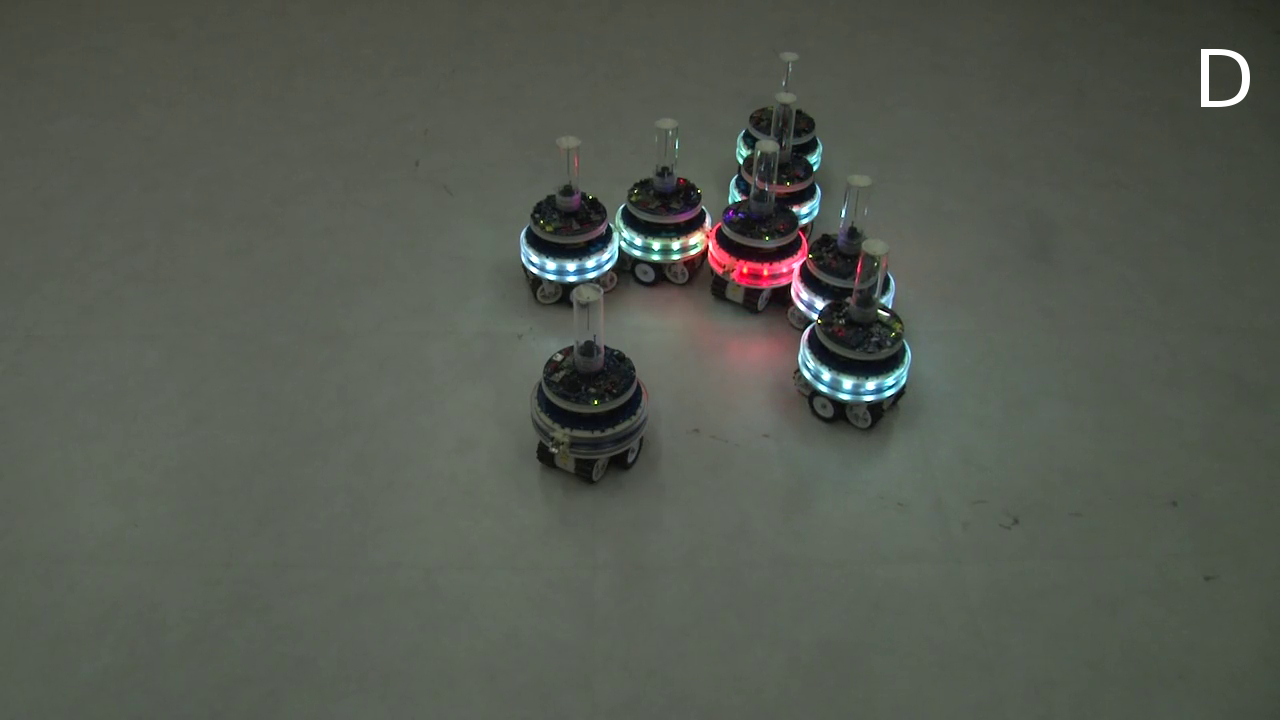}
\caption{\textbf{Fault tolerance: adaptation to a faulty brain.} Brain robots are illuminated in red. (\textbf{A,B}) A composite robotic body suffers from a failure of its brain. The three child robots of the brain node detect the fault.  Each child robot detaches to become the brain of one of three new composite robots. (\textbf{C}, \textbf{D}) The three new composite robots merge to form once again a single composite robot with a single brain.}
\label{fault_tolerance}
\end{figure}

The structural flexibility enabled by \Vcnss\ allows robotic systems to display high levels of fault tolerance, as described below, and enables composite robots to borrow specialized body parts from each other. It would also allow a single robotic system to operate at different scales and to choose between distributed parallel operation and centralized monolithic operation. 
In Fig.~\ref{fault_tolerance}, we show how body splitting and merging capabilities allow a robot to recover even from the loss of its brain. Using the ability to split into new bodies, a child robot can react to a fault in its parent by detaching to create a new independent composite robot of which it will be the brain. In the figure, the programmed behavior is for each of the newly independent composite robots to merge into a shape as close as possible to the original. In this particular case, it was the brain robot that suffered from a fault. A similar mechanism would allow composite robots to excise individual faulty robots from anywhere in their body.

In this report, we have presented the culmination of a research program we started more than ten years ago.  The \Vcns\ technology we developed transforms a collection of self-assembled autonomous robots into a single body controlled by a single brain. A robot equipped with a \Vcns\ no longer has to be built for a particular task. Rather, the size, topology and function of a robot can be decided on the fly by merging or splitting existing robot bodies. 
Future work will involve designing control algorithms for VNS-enabled systems that can determine which body configurations are appropriate under which circumstances. Currently, we pre-program desired body-configurations the robots should autonomously form. However, for our system to reach higher levels of autonomy, it would be desirable for the system itself to choose an appropriate body shape based on its current task and environment.
VNS-enabled machines that can autonomously choose their own body configuration will enable an entirely new class of adaptivity that combines morphological and behavioral responses.

\bibliography{references}

\begin{thebibliography}{1}

\bibitem{Bongard-Science-2006}
Josh Bongard, Victor Zykov, and Hod Lipson.
\newblock {Resilient machines through continuous self-modeling}.
\newblock {\em Science}, 314(5802):1118--1121, November 2006.

\bibitem{ChrOGrDor2007:ram}
A.~L. Christensen, R.~O'Grady, and M.~Dorigo.
\newblock Morphology control in a self-assembling multi-robot system.
\newblock {\em IEEE Robotics \& Automation Magazine}, 14(4):18--25, 2007.

\bibitem{DorFloGam-etal2013:ram}
M.~Dorigo, D.~Floreano, L.~M. Gambardella, F.~Mondada, S.~Nolfi, T.~Baaboura,
  M.~Birattari, M.~Bonani, M.~Brambilla, A.~Brutschy, D.~Burnier, A.~Campo,
  A.~L. Christensen, A.~Decugni{\`e}re, G.~{Di Caro}, F.~Ducatelle,
  E.~Ferrante, A.~F{\"o}rster, J.~Guzzi, V.~Longchamp, S.~Magnenat,
  J.~{Martinez Gonzales}, N.~Mathews, M.~{Montes de Oca}, R.~O'Grady,
  C.~Pinciroli, G.~Pini, P.~R{\'e}tornaz, J.~Roberts, V.~Sperati, T.~Stirling,
  A.~Stranieri, T.~St{\"u}tzle, V.~Trianni, E.~Tuci, A.~E. Turgut, and
  F.~Vaussard.
\newblock Swarmanoid: A novel concept for the study of heterogeneous robotic
  swarms.
\newblock {\em IEEE Robotics \& Automation Magazine}, 20(4):60--71, 2013.

\bibitem{MatChrOGr-etal2011:iros}
N.~Mathews, A.~L. Christensen, R.~O'Grady, P.~R\'{e}tornaz, M.~Bonani,
  F.~Mondada, and M.~Dorigo.
\newblock Enhanced directional self-assembly based on active recruitment and
  guidance.
\newblock In {\em Proceedings of the 2011 IEEE/RSJ International Conference on
  Intelligent Robots and Systems (IROS'11)}, pages 4762--4769. IEEE Computer
  Society Press, Los Alamitos, CA, 2011.

\bibitem{MonGamFlo-etal2005:ram}
F.~Mondada, L.~M. Gambardella, D.~Floreano, S.~Nolfi, J.-L. Deneubourg, and
  M.~Dorigo.
\newblock The cooperation of swarm-bots: Physical interactions in collective
  robotics.
\newblock {\em IEEE Robotics \& Automation Magazine}, 12(2):21--28, 2005.

\bibitem{OGrChrDor2009:tr}
R.~O'Grady, A.~L. Christensen, and M.~Dorigo.
\newblock {SWARMORPH}: Multirobot morphogenesis using directional
  self-assembly.
\newblock {\em IEEE Transactions on Robotics}, 25(3):738--743, 2009.

\bibitem{TriDor2006:biocyb}
V.~Trianni and M.~Dorigo.
\newblock Self-organisation and communication in groups of simulated and
  physical robots.
\newblock {\em Biological Cybernetics}, 95:213--231, 2006.

\end{thebibliography}
\bibliographystyle{plain}

\end{document}